\documentclass[11pt]{article}

\usepackage[preprint]{acl}

\usepackage{newtxtext} 
\usepackage{newtxmath} 
\usepackage{latexsym}

\usepackage[T1]{fontenc}
\usepackage[utf8]{inputenc}

\usepackage{microtype}

\usepackage{inconsolata}
\usepackage{graphicx}
\usepackage{booktabs}
\usepackage{enumitem}
\usepackage{wrapfig}
\usepackage{algpseudocode}
\usepackage[ruled,vlined]{algorithm2e} 
\usepackage{graphicx}
\usepackage{caption}
\usepackage{subcaption}
\usepackage[misc]{ifsym}

\usepackage{microtype}
\usepackage{amsmath}
\usepackage{colortbl}
\usepackage[utf8]{inputenc}
\usepackage[T1]{fontenc}
\definecolor{lightgray}{rgb}{0.9,0.9,0.9}
\usepackage{caption}
\usepackage{subcaption}
\usepackage{setspace}
\usepackage{url}
\usepackage{multirow}
\usepackage{tabularx}
\usepackage{blindtext}
\usepackage{pgfplots}
\usepackage{tikz}
\usetikzlibrary{er,positioning,bayesnet}
\usepackage{makecell}
\usepackage{tipa}
\usepackage{siunitx}
\usepackage{nicefrac}
\usepackage{listings}
\usepackage[raster,skins, most]{tcolorbox} %
\usepackage{xltabular}
\usepackage{adjustbox}
\usepackage{xurl}
\usepackage{rotating}
\usepackage[normalem]{ulem}

\usepackage{amsmath}   
\usepackage{amsfonts}  
\usepackage{bm}        
\usepackage{bbm}

\usepackage{fontawesome}

\usepackage[dvipsnames]{xcolor}
\definecolor{mypink1}{rgb}{0.858, 0.188, 0.478}

\title{
U-Fold: Dynamic Intent-Aware Context Folding for User-Centric Agents
}

\author{Jin Su$^{1,2,3}$\footnotemark[1], Runnan Fang$^{1,2}$\footnotemark[1], Yeqiu Li$^{2}$, Xiaobin Wang$^{2}$, Shihao Cai$^{2}$, 
\\ \textbf{Pengjun Xie$^{2}$, Ningyu Zhang$^{1}$, Fajie Yuan$^{3}$\footnotemark[2]}
\\ Zhejiang University$^1$
\\ Tongyi Lab, Alibaba Group$^2$
\\ Westlake University$^3$
}

\begin{document}
\maketitle

\newcommand\blfootnote[1]{%
\begingroup
\renewcommand\thefootnote{}\footnote{#1}%
\addtocounter{footnote}{-1}%
\endgroup
}

\blfootnote{* Equal contribution. Work done at Tongyi Lab.}
\blfootnote{\dag  Corresponding author.}

\begin{abstract}
Large language model (LLM)-based agents have been successfully deployed in many tool-augmented settings, but their scalability is fundamentally constrained by context length. Existing context-folding methods mitigate this issue by summarizing past interactions, yet they are typically designed for single-query or single-intent scenarios. In more realistic user-centric dialogues, we identify two major failure modes: (i) they irreversibly discard fine-grained constraints and intermediate facts that are crucial for later decisions, and (ii) their summaries fail to track evolving user intent, leading to omissions and erroneous actions. To address these limitations, we propose U-Fold, a dynamic context-folding framework tailored to user-centric tasks. U-Fold retains the full user--agent dialogue and tool-call history but, at each turn, uses two core components to produce an intent-aware, evolving dialogue summary and a compact, task-relevant tool log. Extensive experiments on $\tau$-bench, $\tau^2$-bench, VitaBench, and harder context-inflated settings show that U-Fold consistently outperforms ReAct (achieving a 71.4\% win rate in long-context settings) and prior folding baselines (with improvements of up to 27.0\%), particularly on long, noisy, multi-turn tasks. Our study demonstrates that U-Fold is a promising step toward transferring context-management techniques from single-query benchmarks to realistic user-centric applications.
\end{abstract}
\section{Introduction}

\begin{figure}[!t] 
    \centering
    \scalebox{1}{
    \includegraphics[width=1\linewidth]{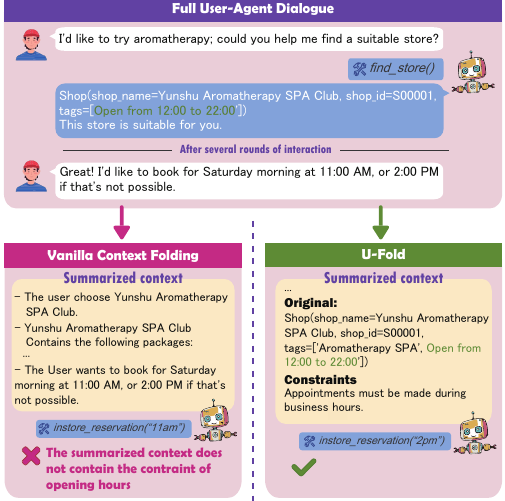} 
    }
    \caption{While vanilla context folding may drop critical constraints after multi-turn interactions and thus lead to incorrect actions, U-Fold preserves fine-grained information through dynamic context folding.}
    \label{fig:advantage}
\vspace{-3mm}
\end{figure}

Large language model (LLM)-based agents have rapidly advanced in tool-augmented applications, from web navigation and software control to life-service assistants~\cite{schick2023toolformer,shen2023hugginggpt,zheng2025deepresearcher,zhou2025prime, wang2023voyager}. A key capability is reasoning over long interaction histories---thoughts, tool calls, and feedback~\cite{yao2022react,wei2022chain,guo2025deepseek, jaech2024openai}. However, naively feeding the entire history to the model leads to context explosion, which exceeds token budgets and overwhelms the model's reasoning ability. This has spurred work on context folding~\cite{wu2025resum,ye2025agentfold,chen2025iterresearch,zhou2025mem1}, which iteratively compresses working memory~\cite{wu2025resum,ye2025agentfold,chen2025iterresearch} to keep context compact while preserving task-critical information.

Although prior folding methods perform well on single-goal long-horizon benchmarks~\cite{wei2025browsecomp,zhou2025browsecomp,mialon2023gaia,shridhar2020alfworld,yao2022webshop}, their behavior in user-centric settings remains underexplored. In realistic practice, an agent interacts with a user over multiple turns, repeatedly uses tools, and must handle user's intent that changes over time~\cite{tau-bench,tau2-bench,qian2025userbench,he2025vitabench,wang2023mint,lu2025toolsandbox}. These settings require higher accuracy and information completeness from folding. Through systematic observation, we find two key limitations of existing methods in user-centric environments. First, static summaries irreversibly drop fine-grained user constraints and intermediate facts (Figure~\ref{fig:advantage}) needed for correct tool use in later turns. Second, they do not track \emph{shifting} user intent, often yielding summaries that lag behind or misrepresent the user's current needs (Figure~\ref{fig:toolcall_cnt}).

To address these challenges, We propose \textbf{U-Fold}, a dynamic, intent-aware context-folding framework for user-centric agents. Instead of static summarization, U-Fold keeps the full user--agent history and, at each turn, builds a compact user-centric working context via two lightweight modules: (i) \textbf{Conversation Summarization}, which tracks dialogue evolution and maintains an up-to-date view of user intent; and (ii) \textbf{Dynamic Data Extraction}, which filters structured tool outputs to retain only fields relevant to current goals. The folded context is then used for reasoning and tool invocation. By aligning context construction with evolving intent, U-Fold reduces redundancy while preserving fine-grained constraints and intermediate facts essential for user-centric tasks.

We evaluate U-Fold on user-centric benchmarks~\cite{tau-bench,tau2-bench,he2025vitabench} and find it consistently outperforms ReAct~\cite{yao2022react} (71.4\% win rate in long-context settings; Figure~\ref{fig:folding_analysis}) and prior folding baselines (up to +27.0\%). Our error analysis further highlights three dominant failure modes: (i) \textbf{\textit{Miscomprehension of User Intent}}, (ii) \textbf{\textit{Omission of Critical User Information}}, and (iii) \textbf{\textit{Unrecognized User Errors}}. U-Fold reduces all three by maintaining an intent-aligned, dynamically folded context (Section~\ref{sec:error_analysis}).

To summarize, our contributions are:
\begin{itemize}
    \item We analyze existing context-folding strategies on realistic user-centric benchmarks and identify failures from intent drift and loss of user-specific constraints.
    \item We introduce U-Fold, which combines conversation summarization with dynamic data extraction to build a compact, informative working context that adapts to evolving user intent.
    \item We evaluate U-Fold across user-centric benchmarks, showing consistent gains over ReAct and prior folding baselines; ablations and error analyses underscore the value of dynamic user-centric folding for robust long-horizon behavior.
\end{itemize}

\begin{figure*}[htbp]
\centering
\includegraphics[width=1.0\textwidth]{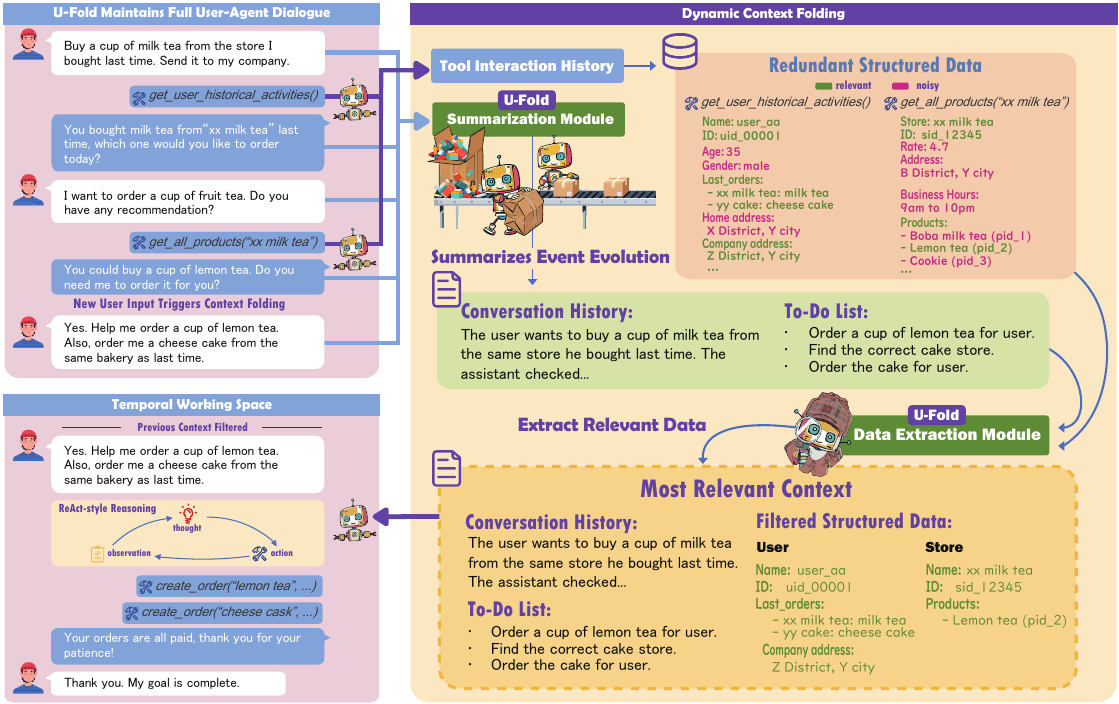}
\caption{
Overview of the U-Fold pipeline. U-Fold consists of two core components for dynamic context folding: (i) a Summarization Module that tracks the evolution of the conversation and maintains an explicit to-do list, and (ii) a Data Extraction Module that filters redundant structured tool outputs and retains only task-relevant information.
}
\label{fig:pipeline}
\vspace{-3mm}
\end{figure*}

\section{Related Work}
\subsection{User-Centric Agent}
Recent research on intelligent agents has increasingly emphasized user-centric interaction~\cite{tau-bench,tau2-bench,qian2025userbench,he2025vitabench,qian2025userrl,wang2023mint}. Traditional evaluations focused on static tasks fail to capture the interactive and dynamic nature of real-world human--agent collaboration. To bridge this gap, $\tau$-bench~\cite{tau-bench} simulates realistic multi-turn dialogues between an agent and a user simulator, incorporating domain-specific API tools and policy constraints. Building on this, $\tau^2$-bench~\cite{tau2-bench} extends the paradigm to dual-control environments, where both the agent and the user can act within a shared environment. In parallel, UserBench~\cite{qian2025userbench} provides a more realistic setting in which agents must proactively clarify underspecified and incrementally revealed user goals. VitaBench~\cite{he2025vitabench} further challenges agents with a complex life-service simulation environment. On the methods side, several works train agents to proactively interact with users~\cite{zhang2025xlam,chen2024learning}, and UserRL~\cite{qian2025userrl} leverages reinforcement learning to explicitly optimize robustness in multi-turn user interactions. However, this line of work largely overlooks systematic management of long, tool-augmented interaction histories. Our framework complements it by providing a dynamic, intent-aware context-folding solution.

\subsection{Context Managment and Compression}
As a standard paradigm, ReAct~\cite{yao2022react} uses no explicit context management: it appends all thoughts, actions, and observations into a continually growing history, leading to context explosion as interaction continues. To mitigate this, one line of work adds external memory (e.g., A-Mem~\cite{xu2025amem}, Mem-OS~\cite{li2025memos}), often at the cost of higher system complexity. Another line compresses the working context~\cite{wu2025resum,chen2025iterresearch,ye2025agentfold,zhou2025mem1,sun2025scaling}. ReSum~\cite{wu2025resum}, AgentFold~\cite{ye2025agentfold}, and IterResearch~\cite{chen2025iterresearch} iteratively summarize prior context to fit token budgets, while~\citet{sun2025scaling} branches temporary contexts for subtasks and later merges them into the main trajectory. However, these compression methods largely target single-goal tasks and do not explicitly support user-centric settings with evolving intent. In contrast, U-Fold offers dynamic, intent-aware folding for user-centric scenarios, maintaining a compact context while handling long tool-augmented histories and shifting user goals.

\section{Methodology}

\subsection{User-Centric LLM-Based Agent Paradigm}
We model the behavior of an LLM-based agent as a Partially Observable Markov Decision Process (POMDP) $(\mathcal{S}, \mathcal{A}, \mathcal{O}, \mathcal{T}, \mathcal{R})$, where $\mathcal{S}$ denotes the state space, $\mathcal{A}$ denotes the action space, $\mathcal{O}$ denotes the observation space, $\mathcal{T}:\mathcal{S}\times\mathcal{A}\rightarrow\mathcal{S}$ denotes the transition function, and $\mathcal{R}$ denotes the reward function. In the user-centric setting, the agent interacts with both the database and the user via a suite of API tools as well as its final response, i.e., $\mathcal{A}=\mathcal{A}_{tool}\cup\mathcal{A}_{resp}$. Accordingly, the state space $\mathcal{S}$ comprises the database state and the user state, $\mathcal{S}=\mathcal{S}_{db}\otimes\mathcal{S}_{user}$, and the observation space $\mathcal{O}$ consists of tool-call outputs and user feedback, $\mathcal{O}=\mathcal{O}_{db}\cup\mathcal{O}_{user}$.

At the outset, the user initiates the conversation by issuing a query $q_1$. The agent then addresses the query by repeatedly executing cycles of \emph{Thought}, \emph{Action}, and \emph{Observation} under the ReAct framework~\cite{yao2022react}. At cycle $t$, the agent generates a reasoning trace $\tau_1^t$ and an action $a_1^t$ conditioned on the preceding context:
\begin{equation*}
\resizebox{0.95\hsize}{!}{$
(\tau_1^t,a_1^t)\sim \pi_{\theta}(\cdot \mid q_1,\tau_1^1,a_1^1,o_1^1,\ldots,\tau_1^{t-1},a_1^{t-1},o_1^{t-1}),
$}
\end{equation*}
where $\pi_\theta$ denotes the agent policy model and $\{o_1^1,o_1^2,\ldots,o_1^{t-1}\}$ denotes the tool-call feedback observed at each cycle in the first turn. The turn terminates when the agent chooses to produce a final response. The resulting sequence of agent decisions can be represented as a trajectory $T_1$:
\begin{equation*}
T_1 = (\tau_1^1,a_1^1,o_1^1,\ldots,\tau_1^t,a_1^t)
\end{equation*}

In the user-centric setting, the user may issue follow-up queries conditioned on the preceding dialogue. At turn $i$, the user query $q_i$ is generated as
\begin{equation}
q_i \sim \pi_{\phi}(\cdot \mid q_1,T_1,q_2,T_2,\ldots,q_{i-1},T_{i-1})
\end{equation}
Here, $\pi_{\phi}$ denotes the user policy that captures user behavior. Likewise, for the agent at turn $i$ and inner cycle $t$, the generation of its reasoning trace $\tau_i^t$ and action $a_i^t$ can be written as
\begin{equation}
\resizebox{0.89\hsize}{!}{$
(\tau_i^t,a_i^t) \sim \pi_{\theta}(\cdot \mid q_1,T_1,\ldots,q_i,\tau_i^1,\ldots,\tau_i^{t-1},a_i^{t-1},o_i^{t-1})
$}
\end{equation}

\subsection{U-Fold: User-Centric Dynamic Context Folding}
Existing context folding methods rely on static, lossy compression, which degrades performance in user-centric scenarios with evolving user intents and fine-grained data requirements. To address this limitation, we propose U-Fold, a dynamic context folding framework. As depicted in Figure~\ref{fig:pipeline}, U-Fold maintains a complete dialogue and tool call information throughout the interaction, while the agent always acts on a compressed view extracted from this full history at each step. U-Fold is triggered on every new user input and decomposes context management into two coordinated components: a \textit{\textbf{Conversation Summarization module}} that maintains an up-to-date, user-centric abstraction of the interaction, and a \textit{\textbf{Dynamic Data Extraction module}} that selectively retrieves task-relevant information from the tool-call history without discarding potentially important details.

\subsubsection{Conversation Summarization}
The Conversation Summarization module is an LLM-based summarizer that compresses the user-agent dialogue into a compact yet informative representation. Given the conversation history, it produces a structured summary that goes beyond recording the agent's past actions by explicitly tracking how the user's goals, constraints, and preferences evolve over time. In addition to the textual summary, the module generates an explicit to-do list that enumerates pending and newly introduced objectives. Each to-do item corresponds to a concrete subtask (e.g., ``verify $X$,'' ``compare $Y$,'' or ``book $Z$ under constraint $C$''), thereby providing an actionable interface between dialogue summarization and downstream planning. Formally, the conversation history $\mathcal{C}_i$ and the resulting summary $\mathcal{M}_i$ are defined as:
\begin{equation}
\mathcal{H}_i = (\tau_i^1,a_i^1,\ldots,\tau_i^t,a_i^t)
\end{equation}
\begin{equation}
\mathcal{C}_i = (q_1,\mathcal{H}_1,q_2,\mathcal{H}_2,\ldots,q_{i-1},\mathcal{H}_{i-1},q_i)
\end{equation}
\begin{equation}
\mathcal{M}_i \sim \pi_{\theta_c}(\cdot \mid \mathcal{C}_i)
\end{equation}
Here $\pi_{\theta_c}$ denotes the policy model of the LLM-based summarizer. By consulting the summary and to-do list, the agent can recover the salient interaction history and infer the current conversational state without revisiting the raw dialogue. This design helps the agent stay aligned with evolving user needs while keeping the context representation concise and tractable.

\definecolor{MyGreen}{HTML}{137E05}

\useunder{\uline}{\ul}{}
\begin{table*}[htbp]
\small
\centering
\renewcommand\arraystretch{1.2}
\resizebox{\textwidth}{!}{%
\begin{tabular}{@{}c|c|cc|ccc|cccc|c@{}}
\toprule
\multirow{2}{*}{\textbf{Model}}         & \multirow{2}{*}{\textbf{Framework}} & \multicolumn{2}{c|}{\textbf{tau-bench}} & \multicolumn{3}{c|}{\textbf{tau2-bench}}      & \multicolumn{4}{c|}{\textbf{VitaBench}}                         & \multirow{2}{*}{\textbf{Improvement}} \\ \cmidrule(lr){3-11}
                                        &                                     & Retail             & Airline            & Retail        & Airline       & Telecom       & Delivery      & In-store      & OTA           & Cross-Scenarios &                                       \\ \midrule
\multirow{4}{*}{Qwen3-4B}               & ReAct                               & \textbf{22.6}      & {\ul 32.0}         & {\ul 21.9}    & {\ul 34.0}    & 19.3          & 9.3           & 5.0           & 0.0           & 0.0             & \textcolor{MyGreen}{$\uparrow$1.1}    \\
                                        & IterResearch                        & 19.1               & 26.0               & 14.1          & 22.0          & \textbf{24.6} & \textbf{13.0} & {\ul 7.0}     & 0.0           & 0.0             & \textcolor{MyGreen}{$\uparrow$3.7}    \\
                                        & ReSum                               & 20.0               & 30.0               & 20.2          & 30.0          & 18.4          & 5.0           & 2.0           & 0.0           & 0.0             & \textcolor{MyGreen}{$\uparrow$3.7}    \\
                                        & \textbf{U-Fold (Ours)}              & {\ul 21.7}         & \textbf{34.0}      & \textbf{22.8} & \textbf{36.0} & {\ul 19.3}    & {\ul 10.0}    & \textbf{8.0}  & 0.0           & \textbf{2.0}    &                                       \\ \midrule
\multirow{4}{*}{Qwen3-Thinking-30B-A3B} & ReAct                               & \textbf{67.0}      & \textbf{46.0}      & {\ul 62.3}    & {\ul 52.0}    & {\ul 30.7}    & {\ul 33.3}    & {\ul 43.5}    & 11.7          & {\ul 15.0}      & \textcolor{MyGreen}{$\uparrow$1.5}    \\
                                        & IterResearch                        & 43.5               & 42.0               & 34.2          & 44.0          & 20.2          & 24.5          & 41.8          & {\ul 12.0}    & 10.8            & \textcolor{MyGreen}{$\uparrow$11.3}   \\
                                        & ReSum                               & 53.9               & 40.0               & 51.8          & 48.0          & \textbf{33.3} & 32.0          & 41.3          & 9.3           & 9.3             & \textcolor{MyGreen}{$\uparrow$6.2}    \\
                                        & \textbf{U-Fold (Ours)}              & {\ul 65.2}         & {\ul 44.0}         & \textbf{70.2} & \textbf{54.0} & 28.9          & \textbf{35.0} & \textbf{44.0} & \textbf{13.8} & \textbf{19.5}   &                                       \\ \midrule
\multirow{4}{*}{DeepSeek-V3.2-Exp}      & ReAct                               & {\ul 70.0}         & {\ul 46.0}         & {\ul 63.2}    & \textbf{70.0} & {\ul 34.2}    & {\ul 48.5}    & {\ul 56.3}    & {\ul 17.5}    & {\ul 20.0}      & \textcolor{MyGreen}{$\uparrow$3.4}    \\
                                        & IterResearch                        & 24.3               & 42.0               & 7.0           & 36.0          & 18.4          & 29.5          & 28.8          & 15.0          & 11.8            & \textcolor{MyGreen}{$\uparrow$27.0}   \\
                                        & ReSum                               & 55.7               & 46.0               & 64.0          & 60.0          & 22.9          & 31.5          & 20.3          & 5.0           & 8.3             & \textcolor{MyGreen}{$\uparrow$15.8}   \\
                                        & \textbf{U-Fold (Ours)}              & \textbf{74.8}      & \textbf{54.0}      & \textbf{64.9} & {\ul 68.0}    & \textbf{37.7} & \textbf{56.8} & \textbf{58.0} & \textbf{21.0} & \textbf{21.0}   &                                       \\ \midrule
\multirow{4}{*}{GPT-4.1}                & ReAct                               & {\ul 72.2}         & {\ul 60.0}         & {\ul 70.2}    & {\ul 64.0}    & {\ul 43.9}    & {\ul 45.5}    & {\ul 55.8}    & {\ul 29.0}    & 17.3            & \textcolor{MyGreen}{$\uparrow$2.5}    \\
                                        & IterResearch                        & 65.2               & 50.0               & 62.5          & 55.0          & 33.3          & 26.0          & 38.5          & 16.0          & 8.8             & \textcolor{MyGreen}{$\uparrow$13.9}   \\
                                        & ReSum                               & 46.1               & 46.0               & 54.2          & 58.5          & 28.1          & 40.0          & 46.5          & 25.8          & {\ul 21.3}      & \textcolor{MyGreen}{$\uparrow$12.7}   \\
                                        & \textbf{U-Fold (Ours)}              & \textbf{73.0}      & \textbf{61.5}      & \textbf{70.8} & \textbf{65.5} & \textbf{44.7} & \textbf{51.3} & \textbf{58.5} & \textbf{33.0} & \textbf{22.3}   &                                       \\ \midrule
\multirow{4}{*}{Claude-4.5-Sonnet}      & ReAct                               & {\ul 85.2}         & \textbf{56.0}      & {\ul 74.1}    & {\ul 66.5}    & {\ul 44.7}    & {\ul 55.8}    & {\ul 56.0}    & {\ul 45.5}    & {\ul 35.5}      & \textcolor{MyGreen}{$\uparrow$2.8}    \\
                                        & IterResearch                        & 76.5               & 52.0               & 71.1          & 63.0          & 29.0          & 50.8          & 59.8          & 30.0          & 27.0            & \textcolor{MyGreen}{$\uparrow$9.5}    \\
                                        & ReSum                               & 69.6               & 54.0               & 70.4          & 65.0          & 42.1          & 53.8          & 55.0          & 34.0          & 35.3            & \textcolor{MyGreen}{$\uparrow$7.3}    \\
                                        & \textbf{U-Fold (Ours)}              & \textbf{86.1}      & {\ul 54.0}         & \textbf{77.2} & \textbf{72.0} & \textbf{45.6} & \textbf{59.0} & \textbf{64.0} & \textbf{48.8} & \textbf{38.0}   &                                       \\ \bottomrule
\end{tabular}%
}
\caption{\textbf{Avg@4 results} on $\tau$-bench, $\tau^2$-Bench, and VitaBench. The best results are highlighted in \textbf{bold}, and the second-best results are \underline{underlined}. The “Improvement” column reports the average performance gain of U-Fold (ours) over each baseline across all domain tasks. We re-implemented and ran all baselines in our unified experimental setup.
\label{tab:main_result}
\vspace{-3mm}
}
\end{table*}

\subsubsection{Dynamic Data Extraction}
Tool calls often return exhaustive user- and task-related information (e.g., full user profiles, product catalogs or complete database records), even though only a small subset is pertinent to the user's current needs. Naively feeding all returned content to the agent quickly bloats the context with noise and distractors. To address this issue, we introduce the Dynamic Data Extraction module, an LLM-based data selector. Conditioned on the history summary and to-do list produced by the Conversation Summarization module, it reviews the full logs of past tool calls and dynamically extracts only the information that is currently useful for resolving the user's pending goals. Irrelevant or redundant fields (e.g., static identifiers, unused attributes, or obsolete options) are pruned, yielding a compact yet semantically sufficient view of the data. The generation of dynamic data $\mathcal{D}_i$ is formally defined as
\begin{equation}
\mathcal{D}_i \sim \pi_{\theta_d}(\cdot \mid \mathcal{M}_i, T_1, T_2,\ldots, T_{i-1}),
\end{equation}
where $\pi_{\theta_d}$ denotes the policy model of the data selector.

\subsubsection{Discussions}
Under the U-Fold framework, long and heterogeneous interaction traces are transformed into a highly condensed yet information-complete context. At each turn, the agent conditions on a compact bundle comprising (i) a user-centric, globally consistent summary of the dialogue trajectory and (ii) a dynamically curated subset of tool-returned data that is strictly relevant to the current goals:
\begin{equation}
\resizebox{0.89\hsize}{!}{$
(\tau_i^t,a_i^t) \sim \pi_{\theta}(\cdot \mid \mathcal{M}_i,\mathcal{D}_i,q_i,\tau_i^1,\ldots,\tau_i^{t-1},a_i^{t-1},o_i^{t-1})
$}
\end{equation}

\begin{figure*}[htbp]
\centering
\includegraphics[width=0.95\textwidth]{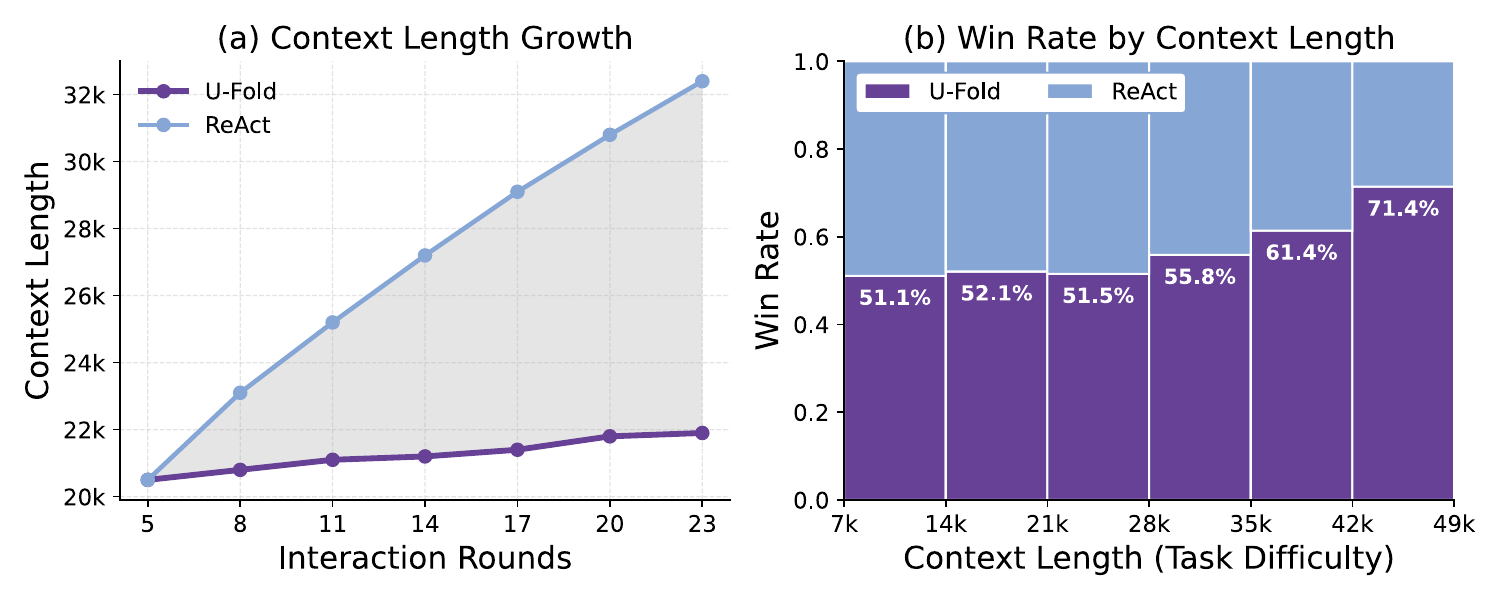}
\caption{
Context folding analysis of U-Fold against ReAct. (a) U-Fold substantially slows context length growth over interaction rounds while preserving task performance. (b) U-Fold win rate over ReAct across bins of final context length (a proxy for task difficulty), where final context length is ReAct's context size at task completion. For each length bin, the win rate is the ratio between the number of tasks solved by U-Fold and those solved by ReAct. U-Fold’s relative advantage increases as final context length grows.
}
\label{fig:folding_analysis}
\vspace{-3mm}
\end{figure*}

This design provides two advantages over standard context folding. First, it enforces \emph{parsimony}: redundant details are aggressively removed, preventing context bloat (see Figure~\ref{fig:folding_analysis} (a)). Second, it maintains \emph{sufficiency}: the dialogue summary and extracted data together retain all information required for correct reasoning and tool use (see Figure~\ref{fig:folding_analysis} (b) and~\ref{fig:toolcall_cnt}). Because both the summary and the extracted data are recomputed after each new user input, U-Fold's compressed context continuously tracks the user's evolving intent throughout the interaction. Consequently, information utilization becomes both simple and efficient: the agent can focus on addressing the current user needs without being distracted by irrelevant historical traces, while still benefiting from the full interaction history maintained in the background.

\section{Experiments}
\subsection{Experimental Setup}
\textbf{Benchmarks.} 
We conduct experiments on several challenging and widely used user-centric benchmarks: $\tau$-bench~\cite{tau-bench}, $\tau^2$-Bench~\cite{tau2-bench}, and VitaBench~\cite{he2025vitabench}. All three benchmarks feature multi-turn interactions with gradually shifting user demands, requiring the agent to invoke tools appropriately to fulfill user requests.

\noindent\textbf{Baselines.} We compare U-Fold with the standard agentic paradigm ReAct~\cite{yao2022react} as well as representative context-folding methods, including ReSum~\cite{wu2025resum} and IterResearch~\cite{chen2025iterresearch}. We evaluate these approaches using both closed-source LLMs (GPT-4.1~\cite{gpt-4.1} and Claude-4.5-Sonnet~\cite{claude-4.5-sonnet}) and open-source LLMs (DeepSeek-V3.2-Exp~\cite{liu2025deepseek}, Qwen3-Thinking-30B-A3B, and Qwen3-4B~\cite{yang2025qwen3}).

\noindent\textbf{Implementation Details.} For U-Fold, we use the same backbone model for both conversation summarization and dynamic data extraction as for the main agent. We use GPT-4.1 as the user simulator. For VitaBench, we use GPT-4.1 as the LLM-based evaluator. During evaluation, we set the temperature to $0.0$ for both the user simulator and the agent. For the main results, we report the average reward over four independent runs (Avg@4). For fair comparison, we re-implemented and ran all baselines in our unified experimental setup.

\subsection{Main Results}
Table~\ref{tab:main_result} displays the comprehensive experimental results. Across all backbones and benchmarks, U-Fold achieves the best performance in nearly every setting and consistently outperforms existing context folding methods, while also rivaling or surpassing the full-context ReAct baseline.

\noindent\textbf{Information-Preserving Compression Beats Full-Context Access.} 
Although ReAct has access to the full dialogue and tool-call history, U-Fold often achieves better performance while operating on a compressed context. Compared with benchmarks with short contexts ($\tau$-bench and $\tau^2$-bench), U-Fold's advantage is more pronounced on VitaBench, where conversations are longer, tool outputs are substantially more verbose, and user intent is more complex. In particular, U-Fold consistently outperforms ReAct across domains such as \textit{Delivery}, \textit{OTA}, and \textit{Cross-Scenarios} (e.g., \textit{DeepSeek-V3.2-Exp Delivery} 56.8 vs.\ 48.5; \textit{GPT-4.1 Cross-Scenarios} 22.3 vs.\ 17.3; \textit{Claude-4.5-Sonnet OTA} 48.8 vs.\ 45.5), indicating that aggressive yet information-preserving compression is especially beneficial in long-horizon, noisy tool-use settings.

The magnitude of the gains also depends on backbone capacity. With a small backbone (Qwen3-4B), improvements over ReAct are modest. We attribute this to the limited ability of smaller models to serve as effective conversation summarizers and data extractors, which constrains the quality of the resulting compressed context (see Section~\ref{sec:transfer}). In contrast, stronger models (DeepSeek-V3.2-Exp, GPT-4.1, and Claude-4.5-Sonnet) yield larger and more consistent improvements, as the U-Fold modules can produce higher-fidelity summaries and more accurate data selection.

\begin{figure}[h]
\centering
\includegraphics[width=1\columnwidth]{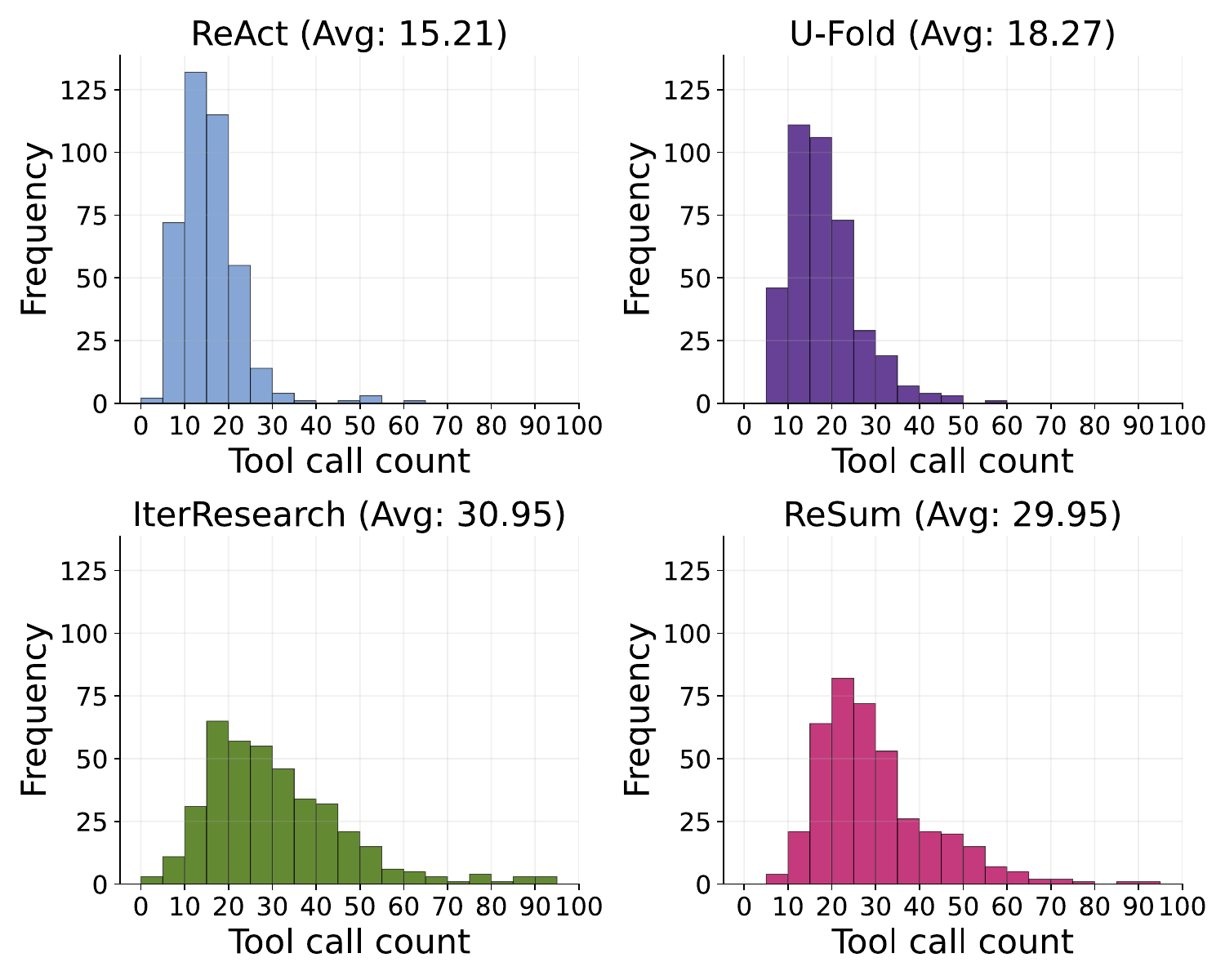}
\caption{
Distribution of tool-call counts under different agentic strategies. Static context folding methods repeatedly invoke tools to recover information lost during the context compression.
}
\label{fig:toolcall_cnt}
\vspace{-3mm}
\end{figure}

\noindent\textbf{Dynamic, Intent-Aware Folding Minimizes Information Loss.}
Compared with IterResearch and ReSum, U-Fold yields substantial and consistent gains across all settings (with improvements of up to 27.0 over IterResearch and 15.8 over ReSum) by addressing two key limitations of static folding methods: (i) the irreversible loss of fine-grained constraints and intermediate facts (Figure~\ref{fig:advantage}), and (ii) insufficient summary content when user intent shifts over time (Figure~\ref{fig:toolcall_cnt}). U-Fold mitigates these issues by retaining the full interaction and tool-call logs and dynamically extracting only the information relevant to the current turn. This design reduces redundant tool calls (Section~\ref{sec:folding_analysis}) and lowers failure rates due to missed user constraints (Section~\ref{sec:error_analysis}), indicating that U-Fold better preserves critical information while still providing the agent with a compact, non-redundant context.

\begin{figure}[htbp]
\includegraphics[width=1\columnwidth]{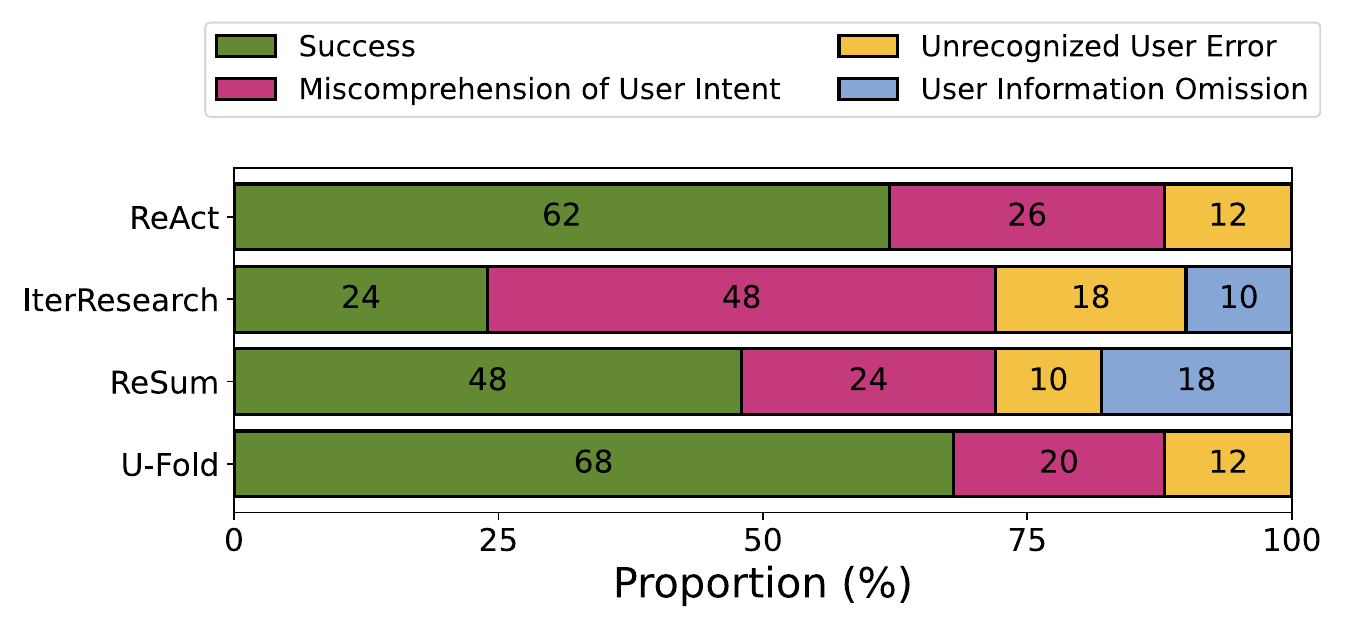}
\caption{
Error analysis on 50 randomly sampled tasks. We report the proportion of successes and three types of failures. U-Fold substantially reduces all error types, especially errors caused by missing user information.
}
\label{fig:error_analysis}
\end{figure}

\subsection{Context Folding Analysis}

\label{sec:folding_analysis}
To assess the context-folding efficiency of U-Fold, we track context-length growth as the number of interaction rounds increases. As shown in Figure~\ref{fig:folding_analysis}(a), U-Fold substantially slows context growth compared with ReAct~\cite{yao2022react}, indicating that our pipeline effectively compresses the working context as the dialogue progresses. Figure~\ref{fig:folding_analysis}(b) further highlights U-Fold's advantage under severe context inflation. Here, the x-axis (\textit{Context Length}) denotes the final context length of ReAct upon task completion, which serves as a proxy for task difficulty. For each interval, we compute the win rate as the ratio of tasks solved by U-Fold to those solved by ReAct. As context length increases, U-Fold's win rate rises steadily, suggesting that our method is particularly beneficial for harder long-context tasks and underscoring the importance of effective context folding.

Figure~\ref{fig:toolcall_cnt} compares the distribution of tool-call counts across different agentic strategies. Conventional context-folding methods (IterResearch~\cite{chen2025iterresearch} and ReSum~\cite{wu2025resum}) periodically summarize and discard the raw history. As user intent evolves, crucial details may no longer appear in the current summary. This information loss forces the agent to repeatedly invoke the same tools to recover missing facts, resulting in substantially more tool calls. In contrast, U-Fold retains the full history and dynamically extracts intent-relevant data whenever the user issues a new query, ensuring that critical information remains accessible in the compressed context. Consequently, U-Fold achieves higher task success while requiring far fewer redundant tool invocations.

\subsection{Error Analysis}
\label{sec:error_analysis}
\begin{figure}[htbp]
\includegraphics[width=1\columnwidth]{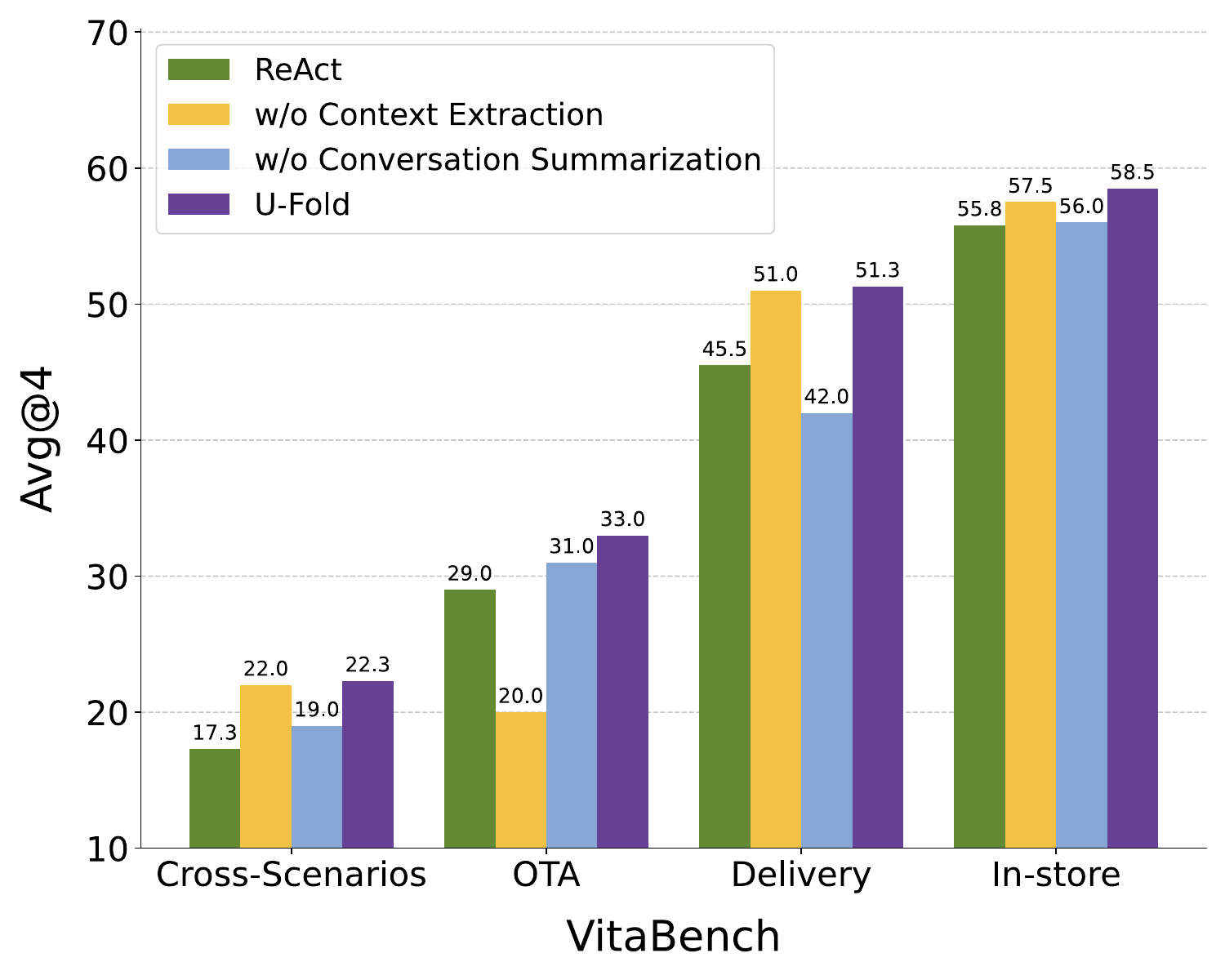}
\caption{
Ablation study of U-Fold. Both modules contribute to the overall gains, and removing either component degrades performance, confirming their complementary roles in effective context folding.
}
\label{fig:ablation}
\end{figure}

\noindent To better understand U-Fold’s behavior on user-centric tasks, we conduct a fine-grained error analysis on 50 randomly sampled VitaBench~\cite{he2025vitabench} tasks. For each method, we manually categorize the failed tasks into three types: (1) \textbf{\textit{Miscomprehension of User Intent}}, where the agent drifts from the user’s evolving goals; (2) \textbf{\textit{Omission of Critical User Information}}, where key constraints or preferences are not reflected in the agent’s context; and (3) \textbf{\textit{Unrecognized User Errors}}, where the agent fails to detect that the user’s inputs are incomplete or inconsistent. As shown in Figure~\ref{fig:error_analysis}, U-Fold achieves the highest success rate and substantially reduces all three error types compared with baselines, with the largest gain in mitigating omissions of critical user information. This aligns with our design: by maintaining full history and dynamically extracting intent-relevant details, U-Fold keeps user constraints explicitly represented in the working context, enabling more faithful intent tracking and more reliable behavior in user-centric, multi-turn interactions.

\subsection{Ablation Study}

\noindent We conduct an ablation study on VitaBench~\cite{he2025vitabench} to quantify the contribution of each U-Fold component. The \textit{w/o Context Extraction} variant retains the full tool-call outputs without filtering, while still using conversation summaries and to-do lists. The \textit{w/o Conversation Summarization} variant, in contrast, keeps the raw dialogue while extracting only the relevant tool logs. As shown in Figure~\ref{fig:ablation}, removing either module degrades performance relative to full U-Fold. The drop is pronounced for \textit{w/o Context Extraction} on the \textit{OTA} domain: \textit{OTA} involves many heterogeneous tools (e.g., tickets, attractions, and transportation options), and without dynamic data extraction the model is exposed to numerous irrelevant candidates, making planning more difficult. Conversely, \textit{w/o Conversation Summarization} performs poorly on \textit{Delivery}, where success depends on tracking user-provided addresses, time windows, and other implicit constraints across multiple turns. Without explicit summarization, the agent often fails to recover these latent requirements from the raw history. Overall, these results suggest that \textbf{Conversation Summarization} and \textbf{Dynamic Context Extraction} are both necessary and complementary to U-Fold's effectiveness.

\begin{table}[]
\small
\centerline
{%
\begin{tabular}{@{}c|cc|cc@{}}
\toprule
\multirow{2}{*}{Framework} & \multicolumn{2}{c|}{In-store} & \multicolumn{2}{c}{Cross-Scenarios} \\ \cmidrule(l){2-5} 
                           & Normal        & Hard          & Normal           & Hard             \\ \midrule
ReAct                      & 43.5          & 11.0          & 15.0             & 4.0              \\
U-Fold                     & \textbf{44.0} & \textbf{27.0} & \textbf{19.5}    & \textbf{12.0}    \\ \bottomrule
\end{tabular}%
}
\caption{
Performance comparison (Avg@4) between ReAct and U-Fold under the standard and hard settings on VitaBench.
}
\vspace{-3mm}
\label{tab:vitabench_hard}
\end{table}
\begin{figure}[t]
\centering
\includegraphics[width=1\columnwidth]{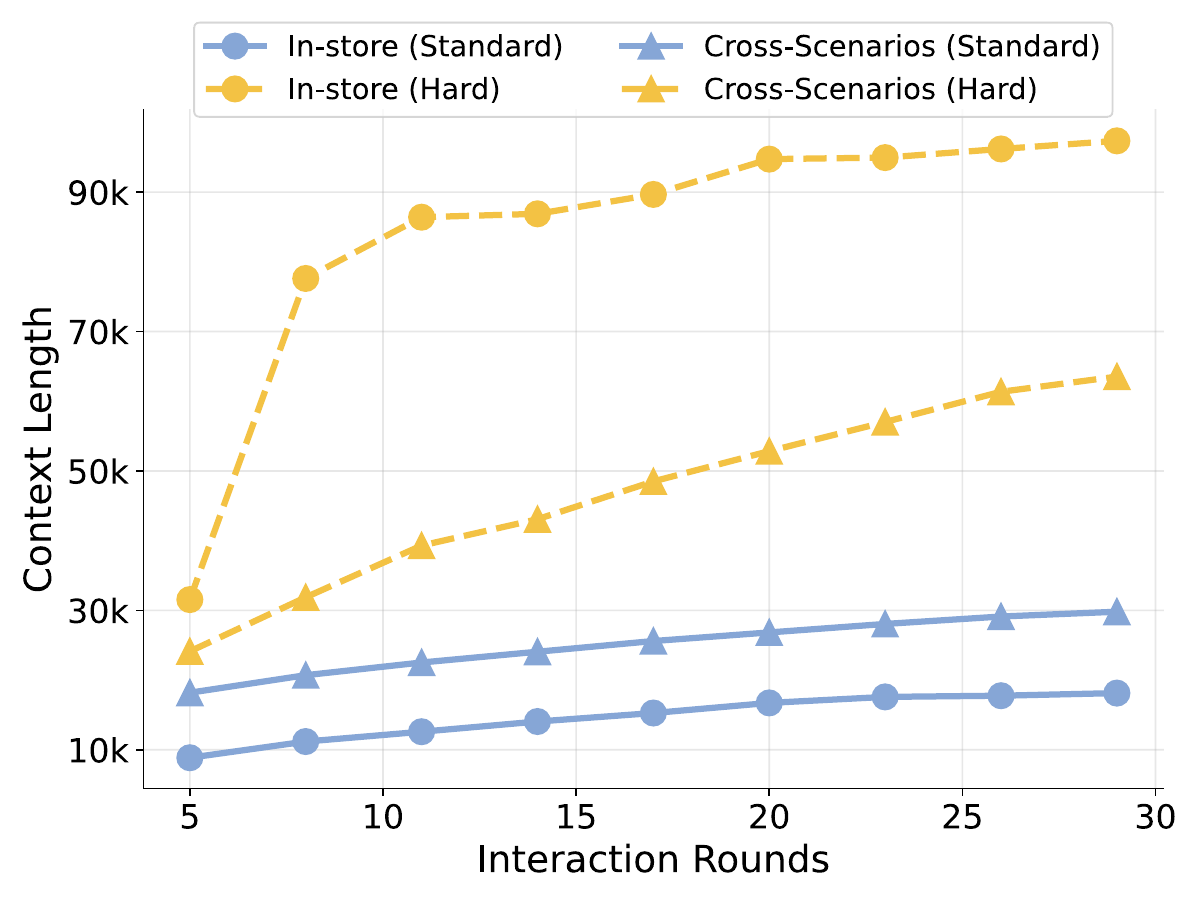}
\caption{
Context-length growth (under the ReAct framework) in the standard and hard settings on VitaBench.
}
\vspace{-3mm}
\label{fig:vitabench_hard}
\end{figure}

\subsection{Further Analysis}
\subsubsection{Evaluation on More Challenging Benchmark}
\noindent To evaluate U-Fold under more severe context inflation, we construct a harder version of VitaBench~\cite{he2025vitabench} for the \textit{In-store} and \textit{Cross-Scenarios} domains using Qwen3-Thinking-30B-A3B~\cite{yang2025qwen3} as the backbone. In this hard setting, tool calls return redundant information as noisy distractors, causing the context length to grow much faster than in the standard setting (Figure~\ref{fig:vitabench_hard}). This setup better reflects real deployments, where tool APIs may expose verbose and noisy payloads.

Table~\ref{tab:vitabench_hard} compares ReAct~\cite{yao2022react} and U-Fold across both difficulty levels. Under the hard setting, U-Fold's advantage becomes substantially larger: performance increases from 11.0 to 27.0 on \textit{In-store} and from 4.0 to 12.0 on \textit{Cross-Scenarios}, despite the much faster context growth induced by redundant tool outputs. These results suggest that U-Fold is particularly effective in the most demanding long-context, user-centric scenarios.

\subsubsection{Capability Transferring through Context Folding}
\begin{table}[]
\large
\resizebox{\columnwidth}{!}
{%
\begin{tabular}{@{}c|cccc@{}}
\toprule
\multirow{2}{*}{Framework} & \multicolumn{4}{c}{VitaBench}                                 \\ \cmidrule(l){2-5} 
                           & Delivery      & In-store     & OTA          & Cross-Scenarios \\ \midrule
ReAct                      & 9.3           & 5.0          & 0.0          & 0.0             \\
U-Fold                     & 10.0          & 8.0          & 0.0          & \textbf{2.0}    \\
U-Fold + Better Folder     & \textbf{17.0} & \textbf{10.0} & \textbf{3.0} & 1.0             \\ \bottomrule
\end{tabular}%
}
\caption{
Capability transfer via U-Fold: Qwen3-4B as the main agent, with U-Fold using either the same backbone or a stronger folder (GPT-4.1). Results are reported as Avg@4 scores.
}
\vspace{-3mm}
\label{tab:transfer}
\end{table}
\label{sec:transfer}
We further examine whether a stronger LLM can act as an informative folder for a weaker agent. In this setting (\textit{U-Fold + Better Folder}), the more capable folder (GPT-4.1~\cite{gpt-4.1}) generates higher-quality conversation summaries and dynamically extracted tool contexts, which are then consumed by the weaker agent (Qwen3-4B~\cite{yang2025qwen3}). As shown in Table~\ref{tab:transfer}, this configuration improves over both ReAct~\cite{yao2022react} and U-Fold, suggesting that capabilities of a large model can be transferred via better context shaping rather than end-to-end inference. Practically, this points to a cost-effective deployment pattern: using a heavyweight LLM sparingly as a folding service to boost cheaper small agents in long-horizon, tool-intensive interactions.
\section{Conclusion}
In this work, we study context-folding strategies for user-centric, tool-augmented settings and reveal key limitations of existing approaches: static summaries cannot reliably track evolving user intent, often discard critical constraints, and consequently induce redundant tool calls or erroneous actions. To address these issues, we propose U-Fold, a dynamic context-folding framework that (i) continuously summarizes the evolving conversation and maintains an explicit to-do list, and (ii) adaptively extracts task-relevant structured information from the full tool-call history. Extensive experiments on challenging user-centric benchmarks, together with comprehensive analyses, show that U-Fold consistently outperforms both ReAct~\cite{yao2022react} and prior folding baselines, especially on long and noisy tasks. These findings suggest that future work on context management should move beyond single-intent summarization toward intent-aware, structure-preserving folding.

\section*{Limitations}
U-Fold delivers substantial gains in user-centric, tool-augmented scenarios, but there remains room for improvement. First, U-Fold currently performs summarization and data extraction at every user turn, rather than explicitly detecting whether user intent has changed. Learning a policy to decide when to trigger folding could further reduce overhead and improve robustness. Second, existing benchmarks still fall short of the complexity of real-world multi-session, multi-user systems. Building more challenging, large-scale benchmarks with richer tool ecosystems and longer, more interdependent tasks is an important direction for future work.

\bibliography{custom}
\clearpage
\appendix

\section{Algorithm Pseudo-Code}

\begin{algorithm}[htbp]
\caption{U-Fold: User-Centric Dynamic Context Folding}
\label{alg:ufold_single_turn}
\DontPrintSemicolon
\KwIn{
New user query $q_t$,
conversation history $C_t=(q_1,\mathcal{H}_1,\dots,q_{t-1},\mathcal{H}_{t-1},q_t)$,
tool history $\{T_1,\dots,T_{t-1}\}$,
agent policy $\pi_{\theta}$,
summarizer $\pi_{\theta_c}$,
data extractor $\pi_{\theta_d}$,
tool set $\mathcal{A}_{tool}$, 
response action $\mathcal{A}_{resp}$
}
\KwOut{Agent trajectory $T_t$}

\BlankLine

$\mathcal{M}_t \sim \pi_{\theta_c}(\cdot \mid C_t)$ \tcp*[l]{Conversation summarization, Eq.~(5)}
$\mathcal{D}_t \sim \pi_{\theta_d}(\cdot \mid \mathcal{M}_t, T_1,\dots,T_{t-1})$ \tcp*[l]{Dynamic data extraction, Eq.~(6)}

\BlankLine
\tcp{Run ReAct-style inner loop conditioned on folded context}
$T_t \leftarrow []$;\;
$i \leftarrow 1$;\;
\While{final response not produced}{
    $(\tau_t^{\,i}, a_t^{\,i}) \sim \pi_{\theta}(\cdot \mid \mathcal{M}_t, \mathcal{D}_t, q_t,
    \tau_t^{\,1:i-1}, a_t^{\,1:i-1}, o_t^{\,1:i-1})$ \tcp*[l]{Eq.~(7)}
    \uIf{$a_t^{\,i} \in \mathcal{A}_{tool}$}{
        $o_t^{\,i} \leftarrow \mathrm{ExecuteTool}(a_t^{\,i})$\;
        $T_t \leftarrow T_t \mathbin{+} [(\tau_t^{\,i}, a_t^{\,i}, o_t^{\,i})]$\;
    }\Else{
        \tcp{$a_t^{\,i} \in \mathcal{A}_{resp}$ is the final natural-language reply}
        $T_t \leftarrow T_t \mathbin{+} [(\tau_t^{\,i}, a_t^{\,i})]$\;
        \textbf{break}\;
    }
    $i \leftarrow i+1$;\;
}
\Return{$T_t$}\;
\end{algorithm}

\section{U-Fold Prompts}

In this section, we provide the full prompts used in U-Fold for all two components: (1) Conversation Summarization, which maintains the evolving summary and to-do list; (2) Dynamic Data Extraction, which filters and structures tool outputs into a compact context. Additionally, we provide the prompt for the Main Agent, which performs ReAct-style reasoning and tool invocation based on the folded context.

\subsection{Conversation Summarization}
The detailed prompt for Conversation Summarization is as follows:

\begin{tcolorbox}[title=Prompt for Conversation Summarization,
    breakable,
    fontupper=\ttfamily\raggedright,
    left=1mm,
    right=1mm,
    top=1mm,
    bottom=1mm
]
You are a dialogue history condenser.

You are given:

- A single, highly condensed, high‑information summary that has been generated from you based on the conversation history.

===

PUT\_HISTORY\_HERE

===

- The new conversation history to tell you what has happened between the user and the assistant.

===

PUT\_CONVERSATION\_HERE

===

Your task:

Based on above information, produce a single, highly condensed, high‑information summary that:

- Preserves all essential facts, constraints, decisions, and assumptions.

- Clearly reflects the chronological order of events and requests.

- Makes the temporal and causal relationships explicit (what happened first, next, and as a result of what).

- Captures the user’s goals, questions, and changes of mind over time.

- Captures the assistant’s key answers, outputs, plans, and action steps (explicitly shows tool name if the tool is used, shows digital identifiers for any data).

- Includes any important time references, deadlines, versions, identifiers, or states (“now”, “later”, “step 2 finished”, etc.).

- If the absolute time is clear, specifically clarify it (day, month etc.) and the relative time (tomorrow, yesterday etc.).

- If the input context is in Chinese, output in Chinese. If the input context is in English, output in English.

Formatting rules:

- Output only one plain text block.

- Do NOT include bullet points, lists, headings, metadata, or commentary about the task.

- Write as a compact narrative, in chronological order, while remaining as concise as possible without losing critical information.

- Output the same language as the input.

Finally, based on the above, output a to-do list that the agent should do to complete the user's goal. Only list those actions have not been done yet:

To-do list (The task should follow the execution order):

    Step1. Task 1
    
    Step2. Task 2
    
    ...

When you generate sub-tasks, detailedly describe the task. Provide all essential facts and constraints to avoid any ambiguity. Don't hallucinate anything not mentioned in the conversation history.
\end{tcolorbox}

\subsection{Dynamic Data Extraction}
The detailed prompt for Dynamic Data Extraction is as follows:

\begin{tcolorbox}[title=Prompt for Dynamic Data Extraction,
    breakable,
    fontupper=\ttfamily\raggedright,
    left=1mm,
    right=1mm,
    top=1mm,
    bottom=1mm
]
You are a context-filtering agent. Your goal is to select all useful past information for the user query.

You are given:

- A list of tools that help you find helpful information for tool use:

===

PUT\_TOOLS\_HERE

===

- A conversation history to tell you what has happened between the user and the assistant:

===

PUT\_CONVERSATION\_HERE

===

- A full interaction history with thought-action-observation triples:

===

PUT\_CONTEXT\_HERE

===

\# Your tasks

Based on the to-do list in conversation history, From the entire thought-action-observation triples, select every OBSERVATION (and only observations) lines that might help answer the new user query, avoid redundant tool calls, or support reasoning.

Important rules:

- You may only extract from **observations** in the full interaction history with thought-action-observation triples. Do not use thoughts or actions.

- You must **not** output any original observation text; only refer to it by line numbers.

- The purpose is context simplification by **line-range selection**, not summarization or rewriting of the original content.

- Prefer observation lines that:

  - Contain facts, intermediate results, or tool outputs relevant to the query’s constraints.
  - Help avoid repeating the same tool call or computation.
  
  - Are likely to be reused for reasoning or final answering.
  
- Irrelevant or weakly related observations should be omitted.

\# Output format

You should output a list of selected context blocks. For each block, you should output:

- Summary: Your concise summary in your own words and how it is related to the to-do list in conversation history (don't extract unnecessary information).

- Original: output the line range in the format "Lines: <start>-<end>". If you only need a single line N, output "Lines: N-N".

- Facts: **ALL** concrete facts explicitly stated in this item (entities, values, states, time info, conditions):
    - For each key information, output its original text. Do NOT rewrite, paraphrase, or edit any extracted text. Every extracted block must be copied verbatim from the original history. Not a single character should be changed, added, or removed.
    
- Constraints: **ALL** limitations, rules, requirements that are explicitly stated or strictly logically implied (no speculation).

- Hint: Generate a detailed, direct guidance for the main task agent to solve the to-do list based on above information. Additionally follow these general principles:

    - When the user refers to past activities, orders, reservations, or interactions (e.g. "the last restaurant", "the hotel I booked before"), do not guess which specific entity they mean.
    
    - Always use tools to retrieve concrete historical records and identify the correct entity by its system identifier (ID or similar). Point out explicitly which tool can be used.
    
    - Never operate on or recommend an entity based only on its name or vague description. The main task agent must first obtain and confirm the entity’s unique system identifier.
    
    - All references to days or times (today, tomorrow, next Monday, this weekend, etc.) must be tied to an explicit absolute date and time.
    
    - For any booking, scheduling, or time-sensitive action, the requested time must lie within the resource’s valid time window (e.g. opening hours, service availability, active period).
    
    - If the available information is not enough to determine whether the requested time is valid or feasible, the main task agent must ask the user for clarification instead of assuming.
    
    - For any operation that requires a quantity or capacity (number of items, tickets, units, rooms, seats, etc.), the main task agent must base it on the user’s explicit request or clearly stated context.
    
    - When the quantity or configuration is not specified and cannot be safely inferred, the main task agent must ask the user directly and must not make up defaults.
    
    - Do not try to change structured attributes (quantity, type, specification, options) indirectly via free-text notes or comments. Structured attributes must be changed using proper fields and supported operations.
    
    - For tasks that involve reordering, modifying, or replacing an existing item, order, or reservation, the main task agent must first retrieve the previous record using tools.
    
    - To modify such records, follow the system’s lifecycle: retrieve → cancel or update as allowed → create a new record if needed. Never assume that the state has changed without confirmation from the tools.
    
    - When the user asks to "order the same as last time" or to "repeat a previous purchase", the main task agent should reuse the concrete identifiers and parameters from the previous record instead of guessing.
    
    - If a place, venue, or address cannot be confidently identified from the provided text, the main task agent must use appropriate tools (e.g., geocoding, map search) to get precise location data (coordinates or canonical address).
    
    - For travel or transport planning, if no direct option is available, the main task agent is allowed to consider indirect routes or nearby locations, but always based on actual tool results, not on speculation.
    
    - When critical information required to complete a task is missing (e.g., date, time, location, number of participants, type of ticket), the main task agent must ask the user explicitly rather than infer or assume values.
    
    - When asking the user questions, avoid leading or suggestive options that could misdirect the user. Ask neutral, direct questions that do not bias the user toward a particular choice.
    
    - If the interaction history or tool results clearly indicate that the user has omitted an essential step (such as purchasing a required ticket, confirming a payment, or finalizing a booking), the main task agent should proactively remind the user and offer to complete that step.
    
    - When entities are described by structured fields or tags, treat them as the ground truth about what exists or is supported.
    
    - Any ability, attribute, or feature not listed in the structured fields should be assumed absent; any attribute listed should be treated as present.
    
    - When answering user questions about those entities, rely directly on the structured fields instead of speculating from general knowledge or typical behavior.
    
    - For any task involving delivery, travel, or time-to-completion, the main task agent must distinguish between start/dispatch time and arrival/completion time.
    
    - Before scheduling a start time, the main task agent should determine the expected duration or time-to-arrival using tools, then choose a start time that satisfies the user’s deadlines or constraints.
    
    - If the user specifies a deadline or required arrival time, the main task agent must explicitly check whether the expected arrival time meets that requirement; if not, it should explain the conflict and propose alternatives.

If nothing is relevant, return an empty output.

If the input context is in Chinese, output in Chinese. If the input context is in English, output in English.
\end{tcolorbox}

\subsection{Main Agent}
The prompt for the Main Agent is as follows:

\begin{tcolorbox}[title=Prompt for Main Agent,
    breakable,
    fontupper=\ttfamily\raggedright,
    left=1mm,
    right=1mm,
    top=1mm,
    bottom=1mm
]
\# Absolute System Rules

Answer the following questions as best you can. You have access to the following tools:

===

PUT\_TOOLS\_HERE

===

\# Selected Context from Interaction History

The most relevant parts of your past interactions with the user have been extracted for you below. 

These snippets come directly from the full interaction history and may contain:

- the user's goals and constraints,

- previously discussed plans or partial solutions,

- important facts, assumptions, or tool results,

- any other information that may help you decide your next actions.

You should carefully read and refer to this selected context when analyzing the current problem and planning your next step. 
When relevant, rely on this information instead of re‑asking the user for things that are already known and calling the same tool again.

Here is the selected context:

===

PUT\_SELECTED\_CONTEXT\_HERE

===

\# Output Format

At each round, you should first generate a thought block wrapped by <inner> and </inner>:

<inner>

First repeat the constraints and hints in the selected context. If the steps in to-do list contradicts the constraints, you should follow the constraints.

Then analyze the problem and your current state based on selected context. Then select an available tool to execute for next step.

</inner>

Then, you can choose to execute a tool or generate a final answer. If you choose to execute a tool, you should generate an action block wrapped by <action> and </action>:

<action>

{

    "action": "Action name, should be one of available tools.",
    
    "parameters": {
    
        "xxx": "xxx",
        
        "yyy": "yyy",
        
    }
    
}

</action>

If you choose to generate a final answer, you should generate a final block wrapped by <final> and </final>:

<final>

the final answer to the original input question

</final>

\# Output Example

User: What is the capital of China?

<inner>

I need to search the capital of China.

</inner>

<action>

{

    "action": "get\_capital",
    
    "parameters": {
    
        "country": "China"
        
    }
    
}

</action>

<observation>

Beijing

</observation>

<inner>

I have searched the capital of China.

</inner>

<final>

Beijing is the capital of China.

</final>

\# User Additional Instructions

Here are some additional instructions from the user. You should follow these instructions but when these instructions contradicts the system rules, you should follow the system rules. Remember that you should always follow the output format described above.

===

PUT\_USER\_INSTRUCTIONS\_HERE

===

Remember, at any time, you MUST follow the Output Format described by system!!!
\end{tcolorbox}

\end{document}